\documentclass[preprint,12pt]{elsarticle}

\usepackage{amssymb}
\usepackage{amsthm}
\usepackage{amsmath}
\usepackage{subfig}

\biboptions{sort&compress}

\journal{Journal of Computational Physics}

\begin{document}

\begin{frontmatter}


\title{Stochastic Physics-Informed Neural Ordinary Differential Equations}

\author[inst1]{Jared O'Leary}
\ead{jared.oleary@berkeley.edu}

\affiliation[inst1]{organization={Department of Chemical and Biomolecular Engineering},
            addressline={University of California, Berkeley}, 
            city={Berkeley},
            postcode={94720}, 
            state={CA},
            country={USA}}

\author[inst2]{Joel A. Paulson}
\ead{paulson.82@osu.edu}

\author[inst1]{Ali Mesbah\corref{cor}}
\ead{mesbah@berkeley.edu}
\cortext[cor]{Corresponding Author}

\affiliation[inst2]{organization={Department of Chemical and Biomolecular Engineering},
            addressline={The Ohio State University}, 
            city={Columbus},
            postcode={43210}, 
            state={OH},
            country={USA}}

\begin{abstract}
Stochastic differential equations (SDEs) are used to describe a wide variety of complex stochastic dynamical systems. Learning the hidden physics within SDEs is crucial for unraveling fundamental understanding of these systems’ stochastic and nonlinear behavior. We propose a flexible and scalable framework for training artificial neural networks to learn constitutive equations that represent hidden physics within SDEs. The proposed stochastic physics-informed neural ordinary differential equation framework (SPINODE) propagates stochasticity through the known structure of the SDE (i.e., the known physics) to yield a set of deterministic ODEs that describe the time evolution of statistical moments of the stochastic states. SPINODE then uses ODE solvers to predict moment trajectories. SPINODE learns neural network representations of the hidden physics by matching the predicted moments to those estimated from data. Recent advances in automatic differentiation and mini-batch gradient descent with adjoint sensitivity are leveraged to establish the unknown parameters of the neural networks. We demonstrate SPINODE on three benchmark \textit{in-silico} case studies and analyze the framework's numerical robustness and stability. SPINODE provides a promising new direction for systematically unraveling the hidden physics of multivariate stochastic dynamical systems with multiplicative noise.

\end{abstract}










\end{frontmatter}

\newcommand*{\citen}[1]{%
  \begingroup
    \romannumeral-`\x 
    \setcitestyle{numbers}%
    \cite{#1}%
  \endgroup   
}


\section{Introduction}

\par Stochastic dynamical systems are ubiquitous in a wide range of science and engineering problems, such as dynamical systems governed by Brownian motion or those that experience random perturbations from their surrounding environment \cite{honerkamp1993stochastic, van1976stochastic, van1992stochastic, volpe2016effective, arnold1974stochastic}. Stochastic differential equations (SDEs) are used to describe the complex behavior of a wide variety of stochastic dynamical systems, including those involving electrical and cell signal processing \cite{schuster2002modelling, pereyra2015survey, young2000stochastic}, colloidal/molecular self-assembly \cite{tang2016optimal, bevan2015controlling}, nucleation processes \cite{singer1953application, penrose2008nucleation}, and predator-prey dynamics \cite{zhu2009competitive, nguyen2017coexistence}. An important challenge in constructing and studying SDEs is that they often contain physics that are either unknown or cannot be directly measured (e.g., free energy and diffusion landscapes \cite{beltran2012colloidal, hummer2005position}, transmission functions in models of disease spread \cite{bain1990applied, korobeinikov2005non}, etc.). Creating a systematic framework to learn the hidden physics within SDEs is thus crucial for unraveling fundamental understanding of stochastic dynamical systems. 

\par A fairly general representation of SDEs is given by:
\begin{equation}
    \label{eqn:gen_SDE}
    dx = f(x,g(x))dt + h(x,g(x))dw, 
\end{equation}
where $x$ is the system state that is generally vector-valued, $t$ is the time, and $w$ is generally a multivariable Gaussian white noise process. The ``modeled'' or ``known'' physics is comprised of $f(\cdot)$, $h(\cdot)$, and the structure of the SDE (i.e., the additive relationship between $f(\cdot)$ and $h(\cdot)$ and the multiplicative relationship between $h$ and $w$). In this work, we consider $g(x)$ to be the ``unmodeled'' or ``unknown'' hidden physics. We thus seek to investigate strategies to create a flexible and scalable framework for systematically learning the hidden physics $g(x)$ within SDEs of form Eq. \eqref{eqn:gen_SDE} from stochastic trajectory data.

\par The most commonly reported methods for learning $g(x)$ from stochastic trajectory data involve evaluating the time limits of the first and second conditional moments \cite{friedrich2000extracting, siegert1998analysis, friedrich2002comment, ragwitz2001indispensable, lamouroux2009kernel, gottschall2008definition, kleinhans2005iterative, gradivsek2000analysis, hegger2009multidimensional, prusseit2008measuring, van2006estimating, kopelevich2005coarse}:
\begin{subequations}
\label{eqn:cond}
\begin{align}
    f(x,g(x)) &= \lim_{\tau \to 0}\frac{1}{\tau} \Big\langle (\xi (t+\tau) - \xi(t)) | \xi(t) = x \Big\rangle  \label{eqn:cond_mean}, \\
    h(x,g(x)) &= \lim_{\tau \to 0}\frac{1}{2\tau} \Big\langle (\xi (t+\tau) - \xi(t))^2 | \xi(t) = x \Big\rangle, \label{eqn:cond_var}
\end{align}
\end{subequations}
where $\xi$ denotes a realization of the stochastic process with a $\delta$-function distribution at the starting point $t$, $\xi(t) = x$, $\tau$ is the sampling time, and the angular brackets denote ensemble averaging. In practice, $\tau \to 0$ must be extrapolated or $\tau$ must be chosen to be sufficiently small to represent the limit. As the lower bound of $\tau$ is often determined by experimental limitations, the primary challenge facing works \cite{ragwitz2001indispensable, lamouroux2009kernel, gottschall2008definition, kleinhans2005iterative, gradivsek2000analysis, hegger2009multidimensional, prusseit2008measuring, van2006estimating, kopelevich2005coarse} is how to determine a robust way to extrapolate $\tau \to 0$. Common approaches to address this challenge involve adding correction terms to Eq. \eqref{eqn:cond_var} \cite{ragwitz2001indispensable}, using autocorrelation functions to simplify Eq. \eqref{eqn:cond_var}  \cite{woolf1994molecular}, employing kernel-based regressions over $\tau$ \cite{lamouroux2009kernel}, and iteratively updating the limit evaluations based on computed probability distributions \cite{kleinhans2005iterative}. However, such methods generally rely on inflexible, data-intensive, and system-specific sampling techniques and/or have been shown to be non-viable when short time linear regions do not exist in the trajectory data \cite{lamouroux2009kernel, hummer2005position, beltran2012colloidal}. 

\par Alternative approaches for learning $g(x)$ leverage Bayesian inference to estimate transition rates along adjacent intervals of $x$, e.g., \cite{hummer2005position, beltran2011smoluchowski, beltran2012colloidal, bevan2015controlling, mittal2008layering, mittal2012pair, ghysels2017position, karimi2018bayesian}. The hidden physics $g(x)$ can then be recovered by exploiting relationships derived from the Fokker-Planck equation \cite{bicout1998electron}. Although Bayesian inference approaches have been shown to be less sensitive to the sampling time than those that depend on extrapolating $\tau \to 0$ \cite{hummer2005position, beltran2012colloidal}, these approaches either (i) learn $g(x)$ at discrete values of $x$ and then fit analytic functions to these discrete values \cite{hummer2005position, beltran2011smoluchowski, beltran2012colloidal, bevan2015controlling, mittal2008layering, mittal2012pair}, or (ii) represent the unknown $g(x)$ using basis functions and learn the coefficients of those basis functions \cite{ghysels2017position, karimi2018bayesian}. The former approach can become intractable when the dimension of $x$ is large, or when $g(x)$ is highly nonlinear and thus requires $x$ to be finely discretized. The latter approach can be highly sensitive to the choice of basis functions and can exhibit other numerical issues. As such, this latter approach often requires \textit{a priori} knowledge about the stochastic system to inform the choice of basis functions \cite{ghysels2017position}.

\par To address the shortcomings described above, we propose a new framework for learning the hidden physics $g(x)$ in Eq. \eqref{eqn:gen_SDE}, which we refer to as stochastic physics-informed neural ordinary differential equations (SPINODE). SPINODE approximates $g(x)$ as an artificial neural network, where the weights and biases within the neural network represent the SDE hidden physics. Artificial neural networks provide a scalable and flexible way of approximating the potentially highly nonlinear relationship between $g(x)$ and continuous values of $x$ without the need for \textit{a priori} assumptions about the form of that relationship \cite{shrestha2019review, zhang2018review, fawaz2019deep}. SPINODE then combines the notions of neural ordinary differential equations (neural ODEs) \cite{chen2018neural, rackauckas2020universal} and physics-informed neural networks (PINN) \cite{raissi2019physics, raissi2020hidden, yang2019adversarial, yang2021b, zhang2019quantifying} to learn the weights and biases within the neural network that approximates $g(x)$ from state trajectory data. If we had access to the true state distribution at particular time points (which is generally non-Gaussian due to the nonlinear terms appearing in Eq. \eqref{eqn:gen_SDE}), we could attempt to identify the neural network parameters that minimize a distributional loss function (e.g., the sum of the Kullback–Leibler divergence between the true and predicted distribution). However, not only would this loss function be more complicated to evaluate, we often do not have direct access to exact state distributions since these must be estimated from a finite set of state trajectories collected from simulations or experiments. Therefore, we opt for a more tractable \textit{moment-matching} framework \cite{hall2005generalized, matyas1999generalized, wooldridge2001applications} in this work, which is an established method in statistics for simplifying the distribution matching problem. There are two key advantages to the moment-matching approach in the context of partially known SDEs:
\begin{itemize}
    \item We only require moments of the state to be measured at discrete time points (with potentially varying sample times) from some known initial state distribution, which are easier to estimate than the full probability distribution or conditional moments. 
    \item The predicted moments of the state based on Eq. \eqref{eqn:gen_SDE} can be estimated using established uncertainty propagation techniques. As long as we can differentiate through the chosen uncertainty propagation method, we can use concepts from the neural ODE framework to compute derivatives needed for efficient training while preserving important features from the underlying SDE. 
\end{itemize}
Although SPINODE can be adapted to handle a variety of different uncertainty propagation methods, we mostly focus on the unscented transform (UT) method \cite{wan2000unscented, julier1997new, sarkka2007unscented} due to its ability to gracefully tradeoff between accuracy and computational efficiency. The UT method, when applied to Eq. \eqref{eqn:gen_SDE}, yields simple analytic expressions for the mean and covariance of the states in terms of the solution to a relatively small set of ODEs. By defining and evaluating the model in terms of ODE solvers, we immediately gain the well-known benefits of such solvers including: (i) memory efficiency, (ii) adaptive computation with error control, and (iii) prediction at arbitrary sets of non-uniform time points \cite{chen2018neural}. All of these benefits are important when developing an efficient training algorithm for the neural network representation of $g(x)$ for which the loss gradient with respect to the neural network parameters can be computed using adjoint sensitivity methods \cite{pontryagin1987mathematical, chen2018neural, rackauckas2020universal}.

\par To highlight the differences between SPINODE and previous methods, let us turn back to Eq. \eqref{eqn:cond}, which essentially computes the time derivative of the mean and covariance of $x$ at some time $t$ by some limit approximation. Previous works \cite{friedrich2000extracting, siegert1998analysis, friedrich2002comment, ragwitz2001indispensable, lamouroux2009kernel, gottschall2008definition, kleinhans2005iterative,  gradivsek2000analysis, hegger2009multidimensional, prusseit2008measuring, van2006estimating, kopelevich2005coarse} have proposed many different strategies for interpolating measured state data from a finite set of discrete time points to estimate this limit; however, these strategies are largely system-specific. SPINODE, on the other hand, uses advanced uncertainty propagation and ODE solvers to directly predict state moment data at any set of time points. Since these underlying methods have been developed to apply to a diverse set of systems, including those that involve high-dimensional, nonlinear, and stiff dynamics, SPINODE can be flexibly applied to systems arising from all different types of applications, which we demonstrate by applying SPINODE to a variety of systems in this work. Furthermore, we note that although SPINODE is primarily described in the context of the first two moments in this paper for simplicity, it can naturally incorporate any number of moments (e.g., skew and kurtosis) when learning $g(x)$. This suggests that SPINODE has the potential to better handle highly non-Gaussian state distributions, which may arise when $f(\cdot)$, $h(\cdot)$, or $g(\cdot)$ are highly nonlinear.

\par We demonstrate the efficacy, flexibility, and scalability of SPINODE on three benchmark \textit{in-silico} case studies. The dynamics of each system are described by SDEs of form Eq. \eqref{eqn:gen_SDE} that contain nonlinear and state-dependent hidden physics terms. The first case study is a two-state model for directed colloidal self-assembly with an exogenous input \cite{tang2013colloidal}, the second is a four-state competitive Lotka-Volterra model with a coexistence equilibrium \cite{xiong2019survival}, and the third is a six-state susceptible-infectious-recovered (SIR) epidemic model for disease spread \cite{cai2017stochastic}. We show that SPINODE is able to efficiently learn the hidden physics within these SDEs with high accuracy. We analyze the numerical robustness and stability of SPINODE and provide suggestions for future research. Furthermore, we have released a fully open-source version of SPINODE on GitHub with end-to-end examples \cite{OLeary2022}, so that interested readers can easily reproduce and extend the results described in this work.

\section{Methods}

\par
\begin{figure*}[ht]
\centering
  \includegraphics[width=\textwidth]{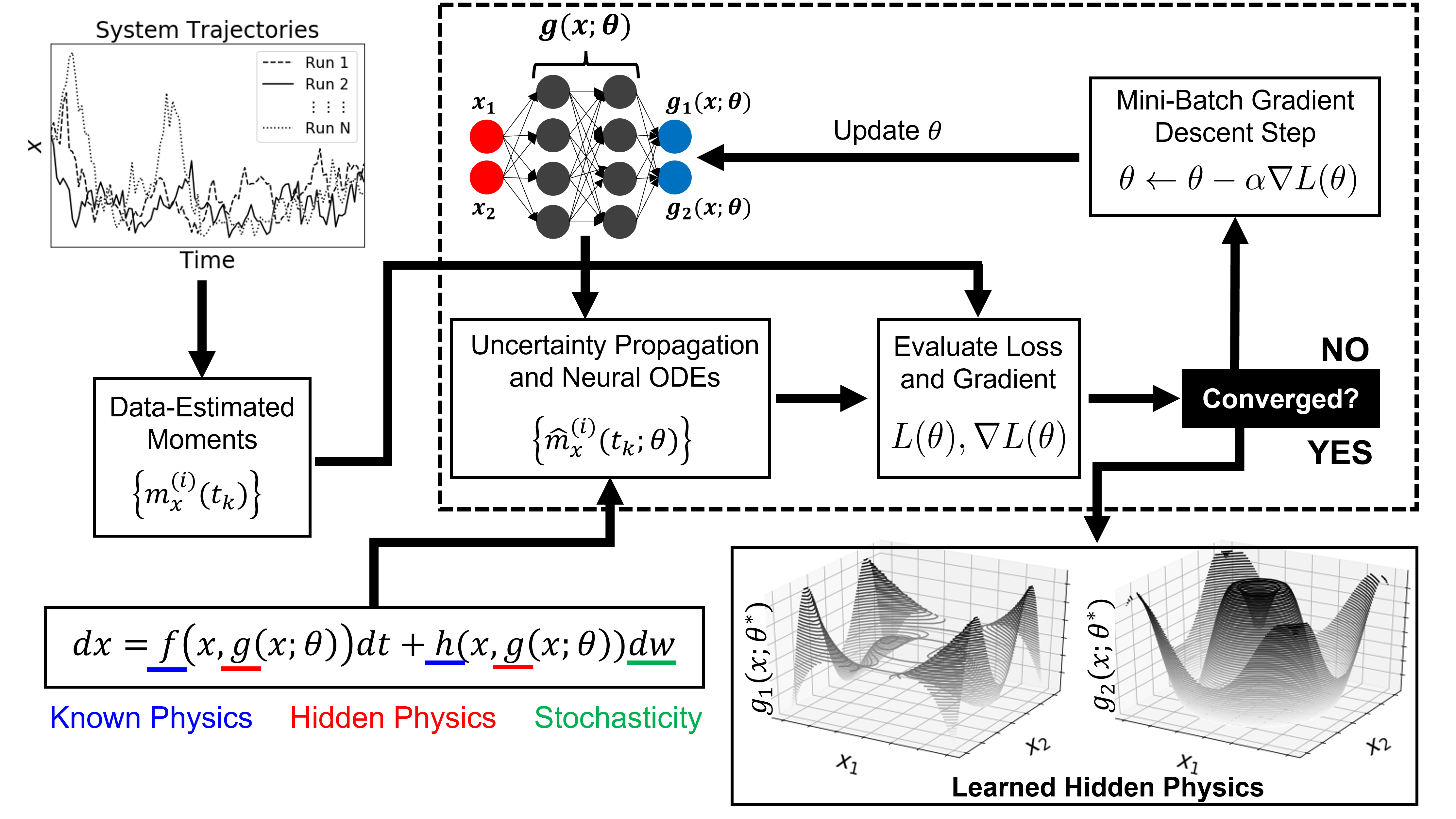}
  \caption{Stochastic physics-informed neural ordinary differential equations framework (SPINODE). The key steps include (i) estimating the time evolution of statistical moments $m^{(i)}_x(t_k)$ from repeated stochastic dynamical system trajectories, (ii) approximating the hidden physics as a neural network (e.g., $g(x;\theta) = [g_1(x;\theta), g_2(x;\theta)]^\top$, where the unknown weights and biases are $\theta$), (iii) using uncertainty propagation to propagate stochasticity through the known structure of the stochastic differential equation, (iv) using ODE solvers within the neural ODE framework to predict the time evolution of statistical moments $\hat{m}^{(i)}_x(t_k)$, and (v) using moment-matching and mini-batch gradient descent with adjoint sensitivity to learn ``optimal'' weights and biases $\theta^*$.
 }
  \label{fgr:SPINODE_Summary}
\end{figure*}

 A schematic overview of the proposed SPINODE method is shown in Fig. \ref{fgr:SPINODE_Summary}. Repeated stochastic dynamical system trajectories are recorded to estimate the time evolution of statistical moments of the stochastic state, $m^{(i)}_x(t_k)$ for all $i=1,\ldots,N_m$ where $N_m$ denotes the total number of moments considered (left). The hidden physics $g(x; \theta)$ are represented by a (deep) neural network that is parameterized by unknown weights and biases denoted by $\theta$ (center). Established uncertainty propagation methods are used to propagate stochasticity through Eq. \eqref{eqn:gen_SDE} and ODE solvers within the neural ODE framework are used to predict the time evolution of the moments for fixed neural network parameters, $\hat{m}^{(i)}_x(t_k; \theta)$ (left to center). A loss function is constructed using the predicted and data-estimated moments (center). Mini-batch gradient descent with adjoint sensitivity is used to update the parameters $\theta$ by minimizing the loss function (right). The hidden physics, $g(x;\theta)$, are considered ``learned'' once the mini-batch gradient descent algorithm converges. The subsequent subsections describe in more detail how data is collected and how SPINODE uses uncertainty propagation, neural ODEs, moment-matching, and mini-batch gradient descent to learn the weights and biases within the neural networks that approximate the unknown hidden physics within SDEs.

\subsection{Data Collection}

\par Data collection is accomplished by repeating stochastic dynamical system trajectories starting from identical initial conditions. Here, $N$ trajectories start from some initial condition $x_0$. During each trajectory, state values $x$ are recorded at time points $t_k$ for $K+1$ total time steps. The $N$ recorded values of each $x(t_k)$ are used to estimate moments $m^{(i)}_x(t_k)$. For simplicity, we primarily focus on the first two moments, the state mean and covariance, which are calculated as follows:
\begin{align}
    &\mu_x(t_k) = \frac{1}{N}\sum_{n=1}^{N}x_n(t_k), \nonumber \\
    &\Sigma_x(t_k) = \frac{1}{N}\sum_{n=1}^{N}(x_n(t_k)-\mu_x(t_k))(x_n(t_k)-\mu_x(t_k))^\top,
    \label{eqn:deMOMENTS}
\end{align}
where $n \in \{1, \ldots, N \}$ denotes the trajectory index.

\par Repeated stochastic trajectories from only one initial condition may not explore a large percentage of the state space. To compensate for this, the stochastic trajectories can be collected from multiple unique initial conditions. In this work, we choose initial conditions by performing a grid search within a specified range of state values of interest. We note, however, that more efficient sampling techniques, e.g., \cite{peeters2001stochastic, reynders2012system, hong2008model, schoukens2019nonlinear, bauer2000numerical} can also be used, which will be explored in future work.

\par Although this work estimates moments of the stochastic states from repeated stochastic trajectories from identical initial conditions, we recognize that this strategy is not applicable for systems in which one does not have control over initial conditions, number of replica runs, or consistent measurement times. In such cases, probability distributions of state trajectories can be learned using methods that may not necessarily require such fine control over the observed trajectory data. Potentially suitable distribution estimation methods include variational autoencoders \cite{kingma2013auto, an2015variational, doersch2016tutorial}, generative adversarial networks \cite{yang2019adversarial, goodfellow2014generative, zoufal2019quantum, mescheder2017adversarial}, and/or energy-based models \cite{lecun2006tutorial, kim2016deep, gustafsson2020energy}. SPINODE is able to accommodate any data collection method from which the shape of the probability distribution (and thus the moments) can be estimated at discrete time points from observed trajectory data. 

\subsection{Moment Prediction}

\par As motivated in the introduction, an important advantage of the moment-matching framework is that we can rely on efficient uncertainty propagation methods that do not require access to the full distribution of the states. Unscented transform (UT) \cite{paulson2017input, wan2000unscented, julier1997new, ponomareva2010new, ebeigbe2021generalized} is one such example of an efficient uncertainty propagation method that estimates moments from a set of well-placed samples (known as \textit{sigma points}) that can be efficiently evaluated using a neural ODE solver. 

Before applying UT to SDEs, let us first summarize the UT method for estimating the moments of a random variable $y = F(z)$ that is some static nonlinear transformation $F : \mathbb{R}^n \to \mathbb{R}^m$ of a random input $z \in \mathbb{R}^n$. We assume knowledge of the mean $m \in \mathbb{R}^n$ and covariance $P \in \mathbb{R}^{n \times n}$ of $z$. Given this information, UT involves the following 3 steps:
\begin{enumerate}
    \item Form the set of $2n + 1$ sigma points from the columns of the matrix $A = \sqrt{(n+\lambda P)}$, which denotes the Cholesky decomposition, as follows
    \begin{align}
        z^{(0)} &= m, \\\notag
        z^{(i)} &= m + \left[ A \right]_i,  &i = 1,\ldots,n, \\\notag
        z^{(i)} &= m - \left[ A \right]_{n-i}, &i = n+1,\ldots,2n,
    \end{align}
    where $[A]_i$ denotes the $i^\text{th}$ column of the matrix $A$. Then, compute the associated weights of each of these sigma points
    \begin{align}
        W_0^{(m)} &= \frac{\lambda}{n + \lambda}, \\\notag
        W_0^{(c)} &= \frac{\lambda}{(n + \lambda) - (1 - \alpha^2 + \beta)}, \\\notag
        W_i^{(m)} &= \frac{1}{2(n + \lambda)}, &i = 1,\ldots,2n, \\\notag
        W_i^{(c)} &= \frac{1}{2(n + \lambda)}, &i = 1,\ldots,2n,
    \end{align}
    where $\lambda$ is a scaling factor defined by
    \begin{align}
        \lambda = \alpha^2 (n + \kappa) - n,
    \end{align}
    and $\alpha$, $\beta$, and $\kappa$ are positive constants. Typically, one should set $\alpha$ to be small (e.g., $10^{-3}$), $\beta=2$, and $\kappa=0$ based on observations from \cite{julier1997new}. 
    \item Transform each of the sigma points as follows
    \begin{align}
        y^{(i)} = F(z^{(i)}),~~~ &i = 0, \ldots, 2n.
    \end{align}
    \item Compute estimates for the mean and covariance of $y$
    \begin{align}
        \hat{\mu}_y &= \sum_{i=0}^{2n} W_i^{(m)} y^{(i)}, \\\notag
        \hat{\Sigma}_y &= \sum_{i=0}^{2n} W_i^{(c)} \left( y^{(i)} - \hat{\mu}_y \right) \left( y^{(i)} - \hat{\mu}_y \right)^\top,
    \end{align}
\end{enumerate}
As shown in X, we can compactly represent UT in matrix form as follows
\begin{align} \label{eq:matrix-ut}
    Z &= \begin{bmatrix}
    m &\cdots &m
    \end{bmatrix} + \sqrt{\alpha^2(n + \kappa)} \begin{bmatrix}
    0 &\sqrt{P} &-\sqrt{P}
    \end{bmatrix}, \\\notag
    Y &= F(Z), \\\notag
    \hat{\mu}_y &= Y w_m, \\\notag
    \hat{\Sigma}_y &= Y W Y^\top, 
\end{align}
where $Z$ denotes the matrix of sigma points and $w_m \in \mathbb{R}^{2n+1}$ and $W \in \mathbb{R}^{2n+1 \times 2n+1}$ are a vector and matrix defined in terms of the mean and covariance weight factors
\begin{align}
    w_m &= [ W_0^{(m)}, \ldots, W_{2n}^{(m)}]^\top, \\\notag
    W &= \left( I - \begin{bmatrix} w_m &\cdots &w_m
    \end{bmatrix} \right) \text{diag}( W_0^{(c)}, \ldots, W_{2n}^{(c)} ) \left( I - \begin{bmatrix} w_m &\cdots &w_m
    \end{bmatrix} \right)^\top
\end{align}
and $I$ denotes the identity matrix of appropriate size. This representation will be helpful when applying UT to SDEs of the form Eq. \eqref{eqn:gen_SDE}. Since both $x$ and $w$ are random quantities, it is more convenient to write out the SDE in the following form
\begin{align} \label{eq:augmented-sys}
    \frac{dz(t)}{dt} = F(z(t) ; \theta) + D e(t),
\end{align}
where $e(t)$ is a zero-mean white noise process with covariance $Q_c(t)$ and $D$ is a dispersion matrix. 
We can express Eq. \eqref{eqn:gen_SDE} in this form by defining an augmented state $z(t) = [x(t), w(t)]^\top$ and defining $F(\cdot)$ and $D$ as follows
\begin{align*}
    F(z(t) ; \theta) &= \begin{bmatrix}
    f(x(t), g(x(t); \theta)) + h(x(t), g(x(t); \theta)) w(t) \\
    0
    \end{bmatrix}, ~
    D &= \begin{bmatrix}
    0 &0 \\
    0 &I
    \end{bmatrix}.
\end{align*}
As shown in \cite{sarkka2007unscented} (Algorithm 4.4), the predicted mean and covariance for any time $t \geq t_k$ can be computed from the initial mean $m(t_k) = [\mu_x(t_k), 0]^\top$ and covariance $P(t_k) = \text{diag}(\Sigma_x(t_k), I)$ (estimated from data as discussed in the previous section) by integrating the following differential equations
\begin{align} \label{eqn:sigma_ode}
    \frac{dm(t)}{dt} &= F(Z(t) ; \theta) w_m, \\\notag
    \frac{dP(t)}{dt} &= Z(t)F^\top(Z(t); \theta) + F(Z(t) ; \theta) W Z^\top(t) + D Q_c(t) D^\top,
\end{align}
where the sigma points $Z(t)$ are defined similarly to that in Eq. \eqref{eq:matrix-ut}, with $m(t)$ and $P(t)$ now being functions of time. We can then recover the original state mean and covariance by simple transformation of the augmented state
\begin{align} \label{eqn:moment_predict}
    \hat{\mu}_x(t | t_k ; \theta) &= \begin{bmatrix} I &0 \end{bmatrix} m(t ; \theta), ~~ \hat{\Sigma}_x(t | t_k ; \theta) = \begin{bmatrix} I &0 \\ 0 &0 \end{bmatrix} P(t ; \theta),
\end{align}
where we have used the notation $t | t_k$ to denote predicted quantities given initial information at time $t_k$. 

\par We use ODE solvers within the neural ODE framework \cite{chen2018neural, rackauckas2020universal} to integrate Eq. \eqref{eqn:sigma_ode} since $F(\cdot)$ is defined in terms of an embedded neural network used to represent the unknown/hidden physics $g(\cdot)$. The flexible choice of ODE solver provides SPINODE with the ability to accurately handle systems with  high-dimensional, stiff, and/or nonlinear dynamics. Another advantage of explicitly integrating the SDE (as opposed to applying a fixed time step discretization) is that we can handle potentially sparse, non-uniform time grids $\{ t_0, t_1, \ldots, t_K \}$. Although we have only exploited information provided by the first two moments of the state distribution, UT can also straightforwardly incorporate higher-order moment data, as described in \cite{ponomareva2010new, ebeigbe2021generalized}. As shown in Section 4.2, incorporating higher-order moments into the prediction scheme can lead to improved performance when learning $g(\cdot)$ due to better placement of the sigma points. 

\subsection{Moment-Matching}

\par Since we represent the hidden physics $g(x ; \theta)$ with a neural network, we need to define a proper loss function to estimate $\theta$. In other words, given a loss function $L(\theta)$, we can translate our goal of ``learning the hidden physics'' into solving the following optimization problem:
\begin{align} \label{eqn:genCOST}
    \theta^\star = \text{argmin}_\theta ~ L(\theta).
\end{align}
A natural loss function for the moment-matching problem is the reconstruction error of the moments, which can be defined as follows
\begin{align}
    L(\theta) = \sum_{k=1}^K \sum_{i=1}^{N_m} \| m^{(i)}_x(t_k) - \hat{m}^{(i)}_x(t_k | t_{k-1}; \theta) \|^2,
\end{align}
which simplifies to the following expression when only the first two moments are considered
\begin{align} \label{eq:twomomentCOST}
    L(\theta) = \sum_{k=1}^K \| \mu_x(t_k) - \hat{\mu}_x(t_k | t_{k-1}; \theta) \|^2 + \| \Sigma_x(t_k) - \hat{\Sigma}_x(t_k | t_{k-1} ; \theta) \|^2,
\end{align}
where $\| \cdot \|^2$ denotes the sum of squared values of all elements in the vector/matrix. We solve Eq. \eqref{eqn:genCOST} via mini-batch gradient descent, which estimates the gradient of the loss function as follows
\begin{align} \label{eq:loss-gradient}
    \nabla_\theta L(\theta) \approx \frac{1}{B} \sum_{k \in \mathcal{B}} \nabla_\theta L_k(\theta),
\end{align}
where $L_k(\theta) = \| \mu_x(t_k) - \hat{\mu}_x(t_k | t_{k-1}; \theta) \|^2 + \| \Sigma_x(t_k) - \hat{\Sigma}_x(t_k | t_{k-1} ; \theta) \|^2$ is the error in the $k^\text{th}$ data point, $B$ is the number of ``mini-batch'' samples, and $\mathcal{B} \subset \{1, \ldots, K \}$ is a set of $B$ randomly drawn indices. We can efficiently evaluate the gradient estimate in Eq. \eqref{eq:loss-gradient} using the adjoint sensitivity method described in \cite{chen2018neural, pontryagin1987mathematical}. Therefore, SPINODE can be easily implemented using open-source deep learning software such as PyTorch \cite{NEURIPS2019_9015} -- we have provided an implementation for the case studies considered in this work on GitHub \cite{OLeary2022}. 

\subsection{Simplified Training Procedure with Approximate Unscented Transform}

\par Based on the UT-based ODEs in Eq. \eqref{eqn:sigma_ode} and the structure of $F(\cdot)$, the evaluation of the mean and covariance are fully coupled, that is, $m(t)$ and $P(t)$ must be simultaneously integrated to evaluate the loss function and its gradient. Since this procedure can be computationally expensive, it is useful to derive alternative approximations that can lead to a simplified training procedure. A particularly important special case of Eq. \eqref{eqn:gen_SDE} is when the hidden physics $g(\cdot)$ are fully separable, i.e., 
\begin{align} \label{eq:separable-SDE}
    dx = f(x, g_1(x ; \theta_1)) + h(x, g_2(x ; \theta_2)) dw,
\end{align}
where $g_1(x ; \theta_1)$ and $g_2(x ; \theta_2)$ denote two completely independent neural networks (each with their own set of local parameters). According to Eq. \eqref{eqn:sigma_ode}, $\theta = \{ \theta_1, \theta_2 \}$ must still be trained simultaneously since the sigma points depend on both the mean and covariance. 

\par To simplify the training process, we present an approximate UT that formulates independent ODEs that describe the time evolution of the transformed sigma points $Y = F(Z)$, where $Z = [Z^x, Z^w]^\top$:
\begin{align}
    \frac{dY}{dt} = f(Z^x(t), g_1(Z^x(t); \theta)) + h(Z^x(t), g(Z^x(t); \theta)) Z^w(t).
    \label{eqn:sigma_ode_no_couple}
\end{align}
The predictions of $Y$ combined with Eq. \eqref{eq:matrix-ut} can be used to predict the mean and covariance. More importantly, since $Z^w(t)$ has a mean of zero and appears in an additive fashion, as long as the weights are chosen in a symmetric fashion, the $h(Z^x(t), g(Z^x(t); \theta)) Z^w(t)$ term will cancel when evaluating the mean of the state. Therefore, in this case, the predicted state mean only depends on $\theta_1$, i.e., $\hat{\mu}_x(t | t_k ; \theta_1)$. Assuming that the predicted state covariance depends weakly on $\theta_1$, we can then separately train $\theta_1$ and $\theta_2$. In particular, we sequentially solve the following two smaller optimization problems:
\begin{align}
    \label{eqn:simpler_cost}
    \theta_1^\star &= \text{argmin}_{\theta_1} \sum_{k=0}^K \| \mu_x(t_k) - \hat{\mu}_x(t_k | t_{k-1}; \theta_1) \|^2, \\\notag
    \theta_2^\star &= \text{argmin}_{\theta_2} \sum_{k=0}^K \| \Sigma_x(t_k) - \hat{\Sigma}_x(t_k | t_{k-1} ; \theta^\star_1, \theta_2) \|^2.
\end{align}
Note that the second optimization problem above is solved using a fixed functional form for the drift term $g_1(x ; \theta_1^\star)$. Although heuristic in nature, this decomposed training strategy greatly reduces the number of parameters that need to be simultaneously considered when evaluating the loss function gradients. Not only does this significantly reduce computational cost, it also limits the search space such that we are less likely to find solutions that result in overfitting. 

\subsection{Validation Criteria using Predicted State Distribution}

\par It is important to note that there can be many values for parameters $\theta$ that result in small or even zero loss function values since moments only provide limited information about the underlying distributions. In other words, even though two different sets of neural network parameters produce the same loss function value, they may result in substantially different predicted state distributions. We can develop a validation test to determine whether or not a given set of optimal parameter values $\theta^\star$ results in accurate state distributions. In particular, we can evaluate the sum of the Kullback–Leibler (KL) divergence \cite{joyce2011kullback} between the measured state distributions $p_{x_k}$ and predicted $\hat{p}_{x_k}(\theta^\star)$ state distributions from a given initial condition over time, i.e., 
\begin{align}
    \text{Validation Error} = \sum_{k=0}^{K_V} \int_\mathcal{X} p_{x_k}(x) \log\left( \frac{p_{x_k}(x)}{\hat{p}_{x_k}(x ; \theta^\star)} \right)dx,
\end{align}
where $K_V$ denotes the number of validation time steps. Note that one can easily modify this definition to include multiple initial conditions and other controlled input values. Since we cannot evaluate either of these distributions exactly, we need to rely on established sample-based probability density function estimation techniques such as kernel density estimation \cite{chen2017tutorial}. We recommend using this validation error criteria to decide if the hidden physics have been learned accurately enough to make reasonable predictions. Whenever the validation error is large, there may be a need to either modify the training strategy, increase the number of moments considered in the loss function, or collect additional data. Due to its simplicity, it is useful to start with the training procedure described in Section 2.4 and, if it does not pass the validation error test described in this section, apply the more detailed coupled training strategy. 

\section{Case Studies}
\par We demonstrate SPINODE on three benchmark \textit{in-silico} case studies from the literature: (i) a two-state model for directed colloidal self-assembly with an exogenous input \cite{tang2013colloidal}, (ii) a four-state competitive Lotka-Volterra model with a coexistence equilibrium \cite{xiong2019survival}, and (iii) a six-state SIR epidemic model for disease spread \cite{cai2017stochastic}. Each of these stochastic dynamical systems can be modeled by Eq. \eqref{eqn:gen_SDE}, and, since the hidden physics are fully separable in each case, Eq. \eqref{eq:separable-SDE}. 

\par State trajectory data is collected by discretizing Eq. \eqref{eqn:gen_SDE} according to an Euler-Maruyama discretization scheme \cite{sancho1982analytical, kloeden1992stochastic}. These discretized SDEs are meant to represent the ``real'' system dynamics. Data-estimated moments $m^{(i)}_x(t_{k})$ are then collected according to the approach described in Section 2.1 and the SPINODE framework outlined in Sections 2.2--2.4 is used to learn (or reconstruct) the hidden physics $g(x)$ from the collected stochastic trajectory data. SPINODE's performance is evaluated by assessing the accuracy of the reconstructed hidden physics. In each case study, moments $m^{(i)}_x(t_{k})$ are calculated from $10^5$ replicates of $50$ time-step state trajectories from $2000$ unique initial conditions (which leads to $10^5$ total moments $m^{(i)}_x(t_{k})$). As mentioned in Section 2.1, the number of initial conditions could very likely be decreased by employing more advanced sampling strategies, but exploring such strategies is beyond the scope of this work. Section 4.2 examines the relationship between the total number of data points and trajectory replicates and the hidden physics reconstruction accuracy.

\subsection{Case Study 1: Directed Colloidal Self-Assembly with an Exogenous Input}

\par The first case study is a two-state model for directed colloidal self-assembly with an exogenous input \cite{tang2013colloidal}. Here, the voltage of an external electric field is adjusted to mediate the two-dimensional self-assembly of silica micro-particles. The system dynamics are modeled according to Eq. \eqref{eqn:gen_SDE}. Denote $x$ as an order parameter that represents crystal structure (i.e., the system state), $u$ as the electric field voltage (i.e., the exogenous input), $K_b$ as Boltzmann's constant, and $T$ as the temperature:
\begin{align}
    & dx = g_1(x,u)dt + \sqrt{2g_2(x,u)}dw, \nonumber \\
    & g_1(x,u) = \frac{d}{dx} \Big(g_2(x,u)\Big) - \frac{d}{dx} \Big(F(x,u)\Big) \frac{g_2(x,u)}{K_b T}, \nonumber \\
    & g_2(x,u) = 4.5 \times 10^{-3} e^{-(x-2.1-0.75u)^2} + 0.5 \times  10^{-3}, \nonumber \\
    & F(x,u) = 10 K_b T(x-2.1-0.75u)^2.
    \label{eqn:csaDYN}
\end{align}
The hidden physics are the drift coefficient, $g_1(x,u)$, and the diffusion coefficient, $g_2(x,u)$. Note that $g_1(x,u)$ is a function of a $g_2(x,u)$ and the free energy landscape $F(x,u)$. This relationship provides an example of how drift and diffusion coefficients can be used to derive other hidden system physics.

\par We chose this case study because the hidden physics are highly nonlinear and depend on an exogenous input. To our knowledge, no previously reported approach for learning SDE hidden physics has explicitly learned $g(x,u)$. Instead, existing approaches typically seek to learn $g(x)$ at discrete values of $u$ and interpolate \cite{tang2016optimal, bevan2015controlling, beltran2011smoluchowski, beltran2012colloidal, tang2013colloidal}. This requires repeating the entire hidden physics learning procedure for many discrete values of $u$ and, thus, demands a trade-off between computational cost and accuracy. SPINODE, on the other hand, directly learns $g(x,u)$ over the entire $(x,u)$ state space.

\subsection{Case Study 2: Competitive Lotka-Volterra with a Coexistence Equilibrium}

\par The second case study is a four-state competitive Lotka-Volterra model with a coexistence equilibrium \cite{xiong2019survival}. The stochastic dynamics are modeled according to Eq. \eqref{eqn:gen_SDE}. Note that $x=[x_1, x_2]^\top$ and $x_i^{\text{eq}}$ are the coexistence equilibrium points:
\begin{align}
    & dx_1 = g_1(x)_1 dt + \sqrt{2g_2(x)_1} dw_1, \nonumber \\
    & dx_2 = g_1(x)_2 dt + \sqrt{2g_2(x)_2} dw_2, \nonumber \\
    & g_1(x)_1 = x_1(1-x_1-k_1 x_2), \nonumber \\
    & g_1(x)_2 = x_2(1-x_2-k_2 x_1), \nonumber \\
    & g_2(x)_1 = x_1(x_2 - x^{\text{eq}}_2), \nonumber \\
    & g_2(x)_2 = x_2(x_1 - x^{\text{eq}}_1), \nonumber \\
    & x^{\text{eq}}_1 = \frac{1-k_1}{1-k_1 k_2}, \quad x^{\text{eq}}_2 = \frac{1-k_2}{1-k_1 k_2}, \nonumber \\
    & k_1 = 0.4, \quad k_2 = 0.5.
    \label{eqn:lveDYN}
\end{align}
The hidden physics are the two-dimensional drift and diffusion coefficients, $g_1(x_1, x_2)$ and $g_2(x_1,x_2)$. As a result, we seek to train two multi-input, multi-output neural networks that approximate the hidden physics. The drift coefficient neural network takes $x_1$ and $x_2$ as input and outputs $g_1(x_1, x_2)_1$ and $g_1(x_1, x_2)_2$. The diffusion coefficient neural network takes $x_1$ and $x_2$ as input and outputs $g_2(x_1, x_2)_1$ and $g_2(x_1, x_2)_2$. 

\par We chose this case study because both the drift and diffusion coefficients are multi-dimensional and nonlinear. We apply SPINODE to this ``more complex'' SDE to demonstrate the framework's scalability. Note that no aspect of the framework was altered from its implementation for the previous case study.

\subsection{Case Study 3: Susceptible-Infectious-Recovered Epidemic Model}

\par The third case study is a six-state SIR epidemic model for disease spread \cite{cai2017stochastic}. The stochastic dynamics are modeled according to Eq. \eqref{eqn:gen_SDE}:
\begin{align}
    & dS = (b - dS - g(S,I) + \gamma R) dt + \sigma_1 S dw_1, \nonumber \\
    & dI = (g(S,I) - (d + \mu + \delta) I) dt + \sigma_2 I dw_2, \nonumber \\
    & dR = (\mu I - (d + \gamma) R) dt + \sigma_3 R dw_3, \nonumber \\
    & g(S,I) = \frac{k S^h I}{S^h + \alpha I^h}, \nonumber \\
    & b=1, \; d=0.1, \; k = 0.2, \; \alpha = 0.5, \; \gamma = 0.01, \; \mu = 0.05, \nonumber \\
    & \delta = 0.01, \; h = 2, \; \sigma_1 = 0.2, \; \sigma_2 = 0.2, \; \sigma_3 = 0.1. \label{eqn:sirDYN}
\end{align}
The hidden physics is the infection transmission rate, $g(S,I)$, which plays a key role in determining disease spread dynamics in many epidemic models \cite{bain1990applied, korobeinikov2005non, cai2017stochastic, bauer2000numerical, hethcote2000mathematics, capasso1978generalization, liu1986influence, yuan2009global}. The form of $g(S,I)$ is widely considered to be unknown, and each of the above-listed references propose different versions of this function. We apply SPINODE to Eq. \eqref{eqn:sirDYN} to learn $g(S,I)$. We chose this case study to demonstrate that SPINODE can not only broadly learn drift and diffusion coefficients but also can learn specific unknown physics terms within complex SDEs.

\par The hidden physics $g(S,I)$ primarily contribute to the deterministic dynamics (i.e., $f(x,g(x)$ in Eq. \eqref{eqn:gen_SDE}) and appears in the time evolution equations for both $S$ and $I$ in Eq. \eqref{eqn:sirDYN}. The resulting loss function used to train $g(S,I;\theta)$ is then given by:
\begin{equation}
    \label{eqn:utCOST_3}
    \min_{\theta} \sum_{k=0}^K \big \| \hat{\mu}_S(t_k) - \mu_S(t_k)) \big \|^2 + \big \| \hat{\mu}_I(t_k) - \mu_I(t_k)) \big \|^2,
\end{equation}
while the loss functions used to train $g_1(x,u)$ and $g_1(x_1, x_2)$ in the previous two case studies were given by Eq. \eqref{eqn:simpler_cost}.

\section{Results and Discussion}

\begin{figure*}[ht!]
\centering
  \includegraphics[width=\textwidth]{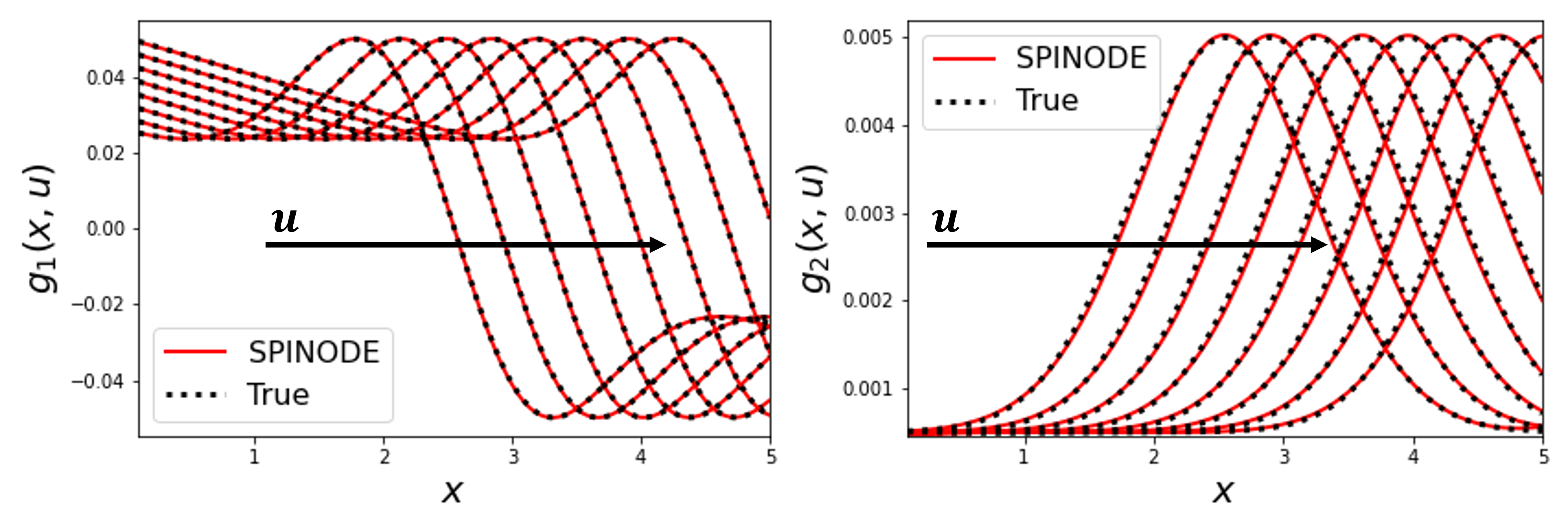}
  \caption{Learned hidden physics of directed colloidal self-assembly system with an exogenous input. SPINODE learns the drift and diffusion coefficients $g_1(x,u)$ and $g_2(x,u)$ of the stochastic dynamical system described by Eq. \eqref{eqn:csaDYN} with high accuracy.
 }
  \label{fgr:csa_rec}
\end{figure*}

\begin{figure*}[ht!]
\centering
  \includegraphics[width=\textwidth]{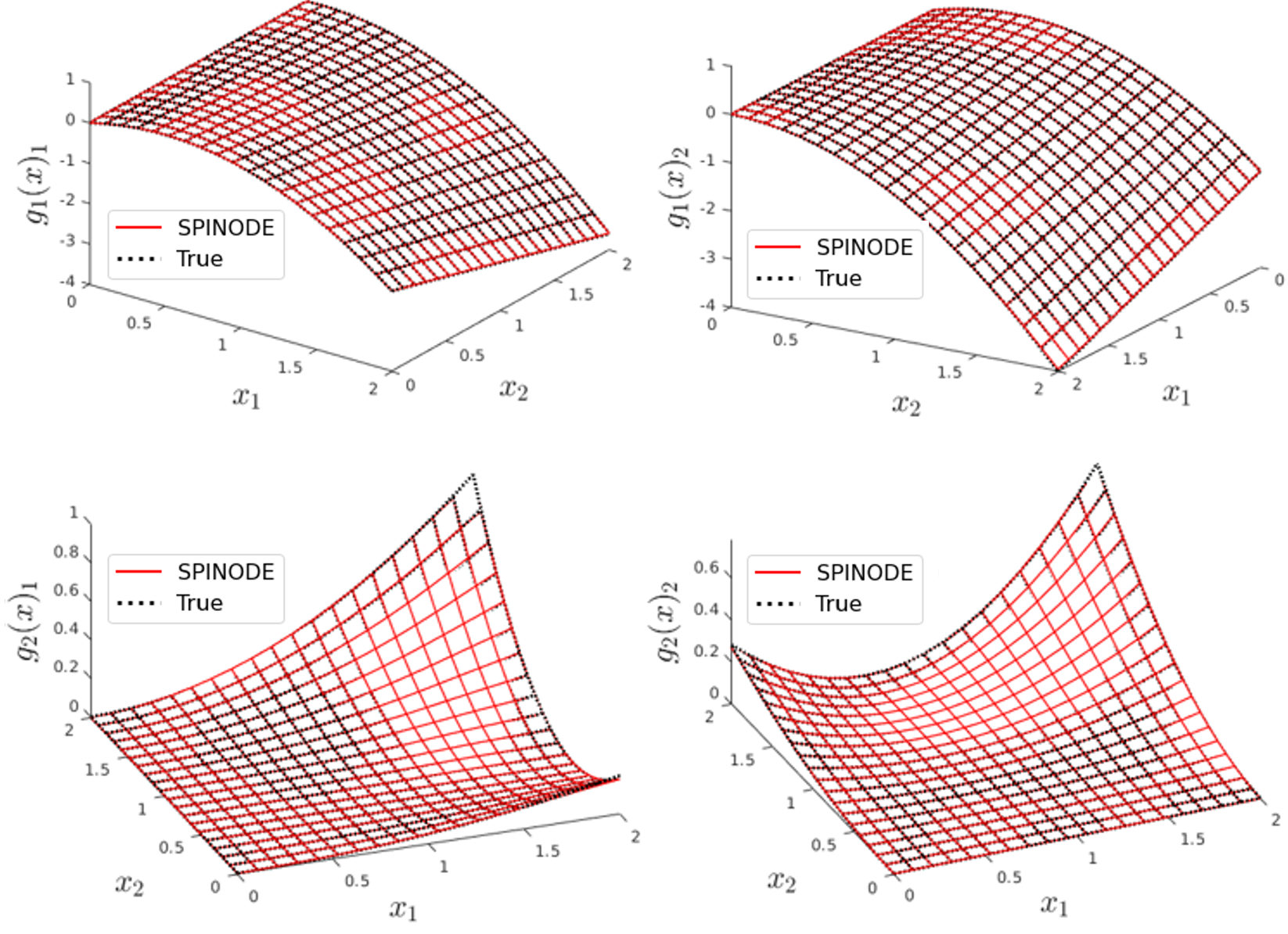}
  \caption{Learned hidden physics of competitive Lotka-Volterra with a coexistence equilibrium. SPINODE learns the drift and diffusion coefficients $g_1(x_1,x_2)_{1,2}$ and $g_2(x_1,x_2)_{1,2}$ of the stochastic dynamical system described by Eq. \eqref{eqn:lveDYN} with high accuracy.
 }
  \label{fgr:lve_rec}
\end{figure*}

\begin{figure*}[ht!]
\centering
  \includegraphics[width=0.6\textwidth]{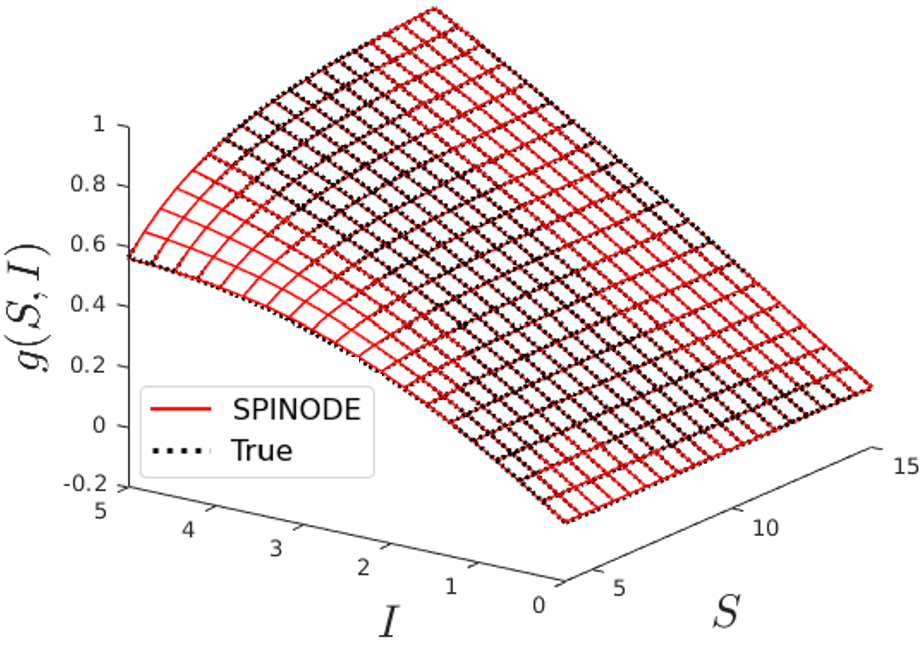}
  \caption{Learned hidden physics of susceptible-infectious-recovered (SIR) epidemic model. SPINODE learns $g(S,I)$ from Eq. \eqref{eqn:sirDYN} with high accuracy.
 }
  \label{fgr:sir_rec}
\end{figure*}

\subsection{Learning Hidden Physics}

\par We demonstrate SPINODE on the case studies outlined in Section 3. In each case study, moments $m^{(i)}_x(t_{k})$ (e.g., means and covariances) are estimated from stochastic trajectory data. The approximate UT method described in Section 2.4 is then used to yield deterministic ODEs that describe the time evolution of the sigma points. An Euler ODE scheme is used to solve these ODEs and thus predict the time evolution of the means and covariances. Mini-batch gradient descent with adjoint sensitivity is then used to (i) match the predicted means and covariances to the data-estimated means and covariances and (ii) train the neural networks $g(x;\theta)$ that approximate the true hidden physics $g(x)$. Note that the time intervals over which the means and covariances are predicted (i.e., the sampling times) are approximately 1/50\textsuperscript{th} of the time it takes each system to reach steady state.

\par
\begin{figure*}[ht!]
\centering
  \includegraphics[width=0.6\textwidth]{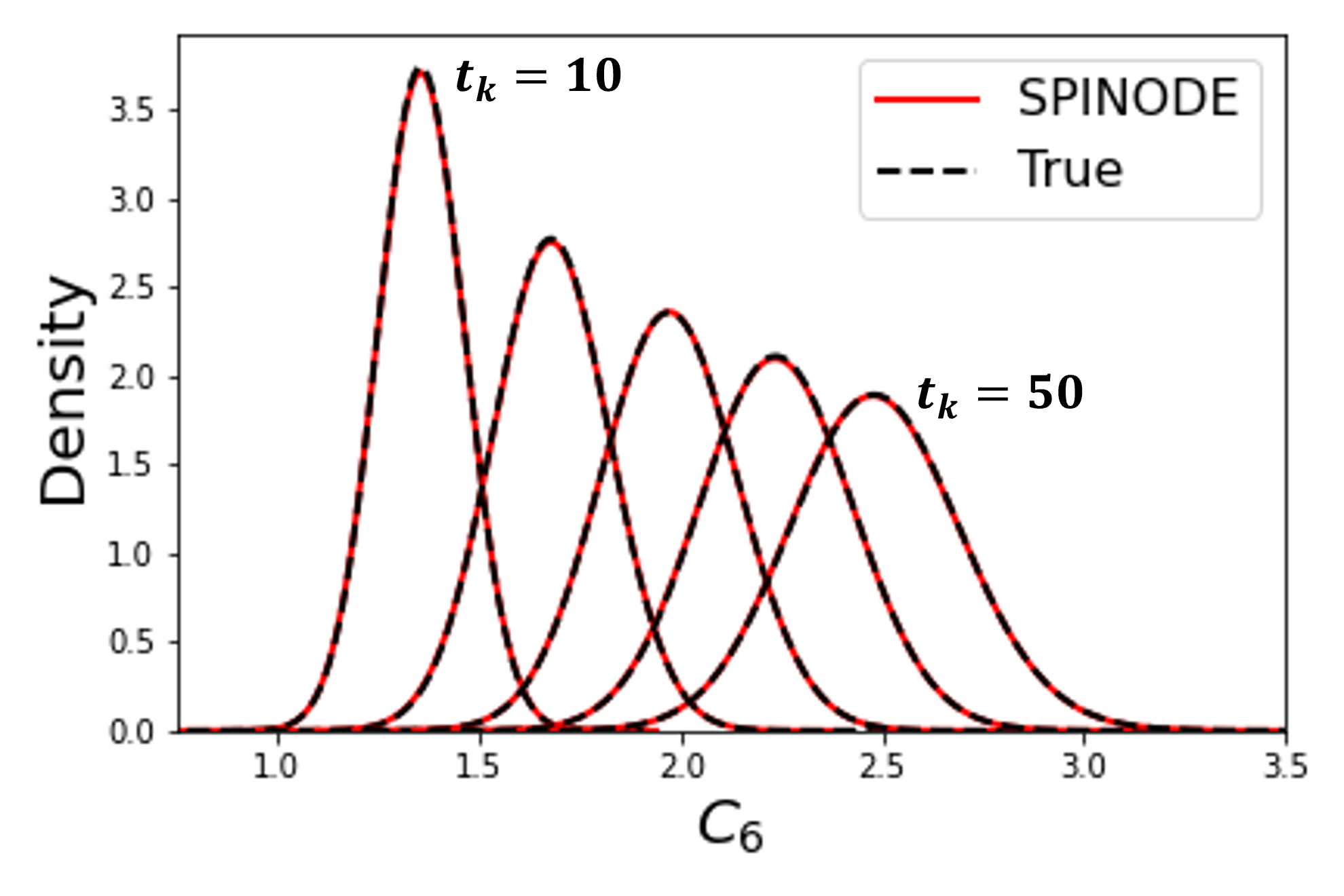}
  \caption{Time evolution of kernel density estimates for directed colloidal self-assembly system with an exogenous input. Trained neural networks $g_1(x,u;\theta_1^\star)$ and $g_2(x,u;\theta_2^\star)$ are used to simulate the system dynamics from a randomly selected initial condition with a randomly selected exogenous input. The true dynamics are then simulated using the same initial condition and exogenous input. In each case, the stochastic trajectory is repeated $10^5$ times and kernel density functions are calculated at each sampling time. Estimates of the kernel density function for the ``true'' and ``learned'' dynamics at select sampling times are plotted against one another. SPINODE reproduces the kernel density function with high accuracy.
 }
  \label{fgr:csa_distcomp}
\end{figure*}

\begin{figure*}[ht!]
\centering
  \includegraphics[width=\textwidth]{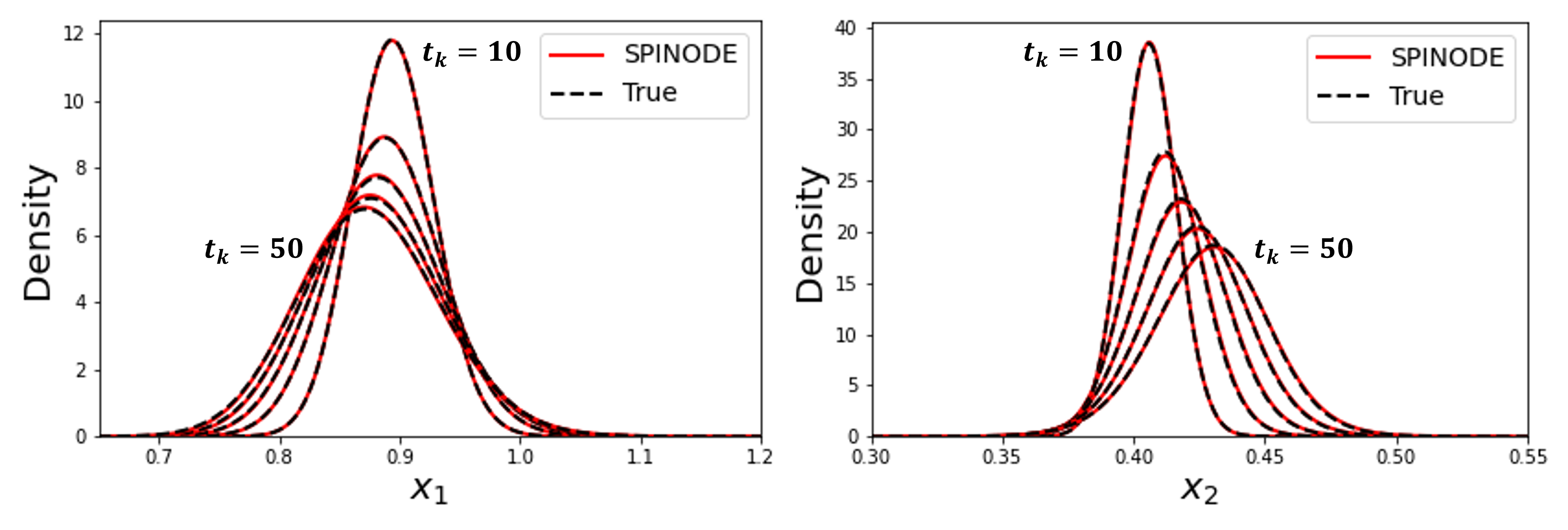}
  \caption{Time evolution of kernel density estimates for Lotka-Volterra with a coexistence equilibrium. Trained neural networks $g_1(x_1,x_2;\theta_1^\star)$ and $g_2(x_1,x_2;\theta_1^\star)$ are used to simulate the system dynamics from a randomly selected initial condition. The true dynamics are then simulated using the same initial condition. In each case, the stochastic trajectory is repeated $10^5$ times and kernel density functions are calculated at each sampling time. Estimates of the kernel density function for the ``true'' and ``learned'' dynamics at select sampling times are plotted against one another. SPINODE reproduces the kernel density function with high accuracy.
 }
  \label{fgr:lve_distcomp}
\end{figure*}

 In principle, SPINODE's performance can vary from run-to-run due to the randomness involved in neural network weight initialization and assigning data-estimated moments $m^{(i)}_x(t_{k})$ to training, validation, and test sets. We thus assessed SPINODE's performance by calculating the root mean squared errors (RMSE) between the learned hidden physics $g(x;\theta^*)$ and the actual system hidden physics $g(x)$ over 30 SPINODE runs with randomly selected initial weight values and training/validation/test set data assignments. Table 1 shows the mean and standard deviations of these RMSEs while Figs. \ref{fgr:csa_rec} -- \ref{fgr:sir_rec} show a visual comparison of $g(x;\theta^*)$ and $g(x)$ for representative runs. In each case, SPINODE learns the hidden physics $g(x;\theta^*)$ with high accuracy and low run-to-run variation. We note that in the real-world, the actual values of the hidden physics $g(x)$ will be unavailable. In these cases, SPINODE's performance should be validated via the methodology described in Section 2.5., i.e., by comparing the data-estimated moments of trajectories from the real dynamics $m^{(i)}_x(t_{k})$ to those generated from the learned dynamics involving $f(\cdot)$, $h(\cdot)$, and $g(x,\theta^*)$. Visual representations of the time-evolution of the probability distributions of the states from randomly selected initial conditions and exogenous input values for the colloidal self-assembly and Lotka-Volterra case studies are shown in Figs. \ref{fgr:csa_distcomp} -- \ref{fgr:lve_distcomp}.

\begin{table}[ht!]
\small
  \label{tbl:rmseSUM}
  \begin{tabular*}{0.95\textwidth}{@{\extracolsep{\fill}}lll}
    \hline
    Case Study  &  RMSE Mean & RMSE Std\\
    \hline
    Colloidal Self-Assembly, $g_1(x,u)$  & $1.33 \times 10^{-4}$ & $2.77 \times 10^{-5}$\\
    Colloidal Self-Assembly, $g_2(x,u)$ & $4.97 \times 10^{-5}$ & $4.67 \times 10^{-6}$\\
    Lotka-Volterra , $g_1(x_1, x_2)_1$ & $9.20 \times 10^{-4}$   & $1.24 \times 10^{-4}$\\
    Lotka-Volterra , $g_1(x_1, x_2)_2$ & $7.91 \times 10^{-4}$  & $7.59 \times 10^{-5}$\\
    Lotka-Volterra , $g_2(x_1, x_2)_1$ & $3.93 \times 10^{-3}$  & $5.97 \times 10^{-5}$\\
    Lotka-Volterra , $g_2(x_1, x_2)_2$ & $4.88 \times 10^{-3}$  & $7.15 \times 10^{-5}$\\
    Susceptible-Infectious-Recovered  , $g(S,I)$ & $2.62 \times 10^{-3}$  & $1.89 \times 10^{-4}$\\
    \hline
  \end{tabular*}
  \caption{Reconstruction root mean square errors of learned hidden physics. SPINODE is used to learn the hidden physics of the case studies in Sections 3.1 (directed colloidal self-assembly with an exogenous input), 3.2 (competitive Lotka-Volterra with a coexistence equilibrium), and 3.3 (susceptible-infectious-recovered epidemic model). The root mean square error (RMSE) between the learned hidden physics $g(x;\theta^*)$ and actual hidden physics $g(x)$ is then calculated. This process is repeated 30 times with randomly selected initial weight values and training/validation/test set data assignments. The means and standard deviations (std) of the calculated RMSEs are shown. For each case study, SPINODE learns the hidden physics $g(x;\theta^*)$ with high accuracy and low run-to-run variation.}
\end{table}

\par We further note that the hidden physics reconstructions shown in Figs. \ref{fgr:csa_rec} -- \ref{fgr:sir_rec} and Table 1 essentially occur under ``ideal'' conditions -- as $g(x;\theta)$ is trained using a large number of moments $m^{(ij)}_x(t_{k})$ that are estimated from a large number of repeated trajectories from a large number of initial conditions (see Section 2.1 and Section 3 for details). In addition, the sampling times are identical to the discretization times used in the Euler-Maruyama simulations that represent the ``true'' system dynamics. This last point motivated the use of an Euler ODE solver to predict the moment time-evolution. In the next section, we assess the performance of SPINODE in terms of decreasing the number of repeated trajectories used to estimate the moments $m^{(ij)}_x(t_{k})$, decreasing the total number of moments $m^{(ij)}_x(t_{k})$ used to train $g(x;\theta)$, altering the uncertainty propagation strategy, and adjusting the sampling time.

\subsection{Numerical Robustness}

\par
\begin{figure*}[ht!]
\centering
  \includegraphics[width=\textwidth]{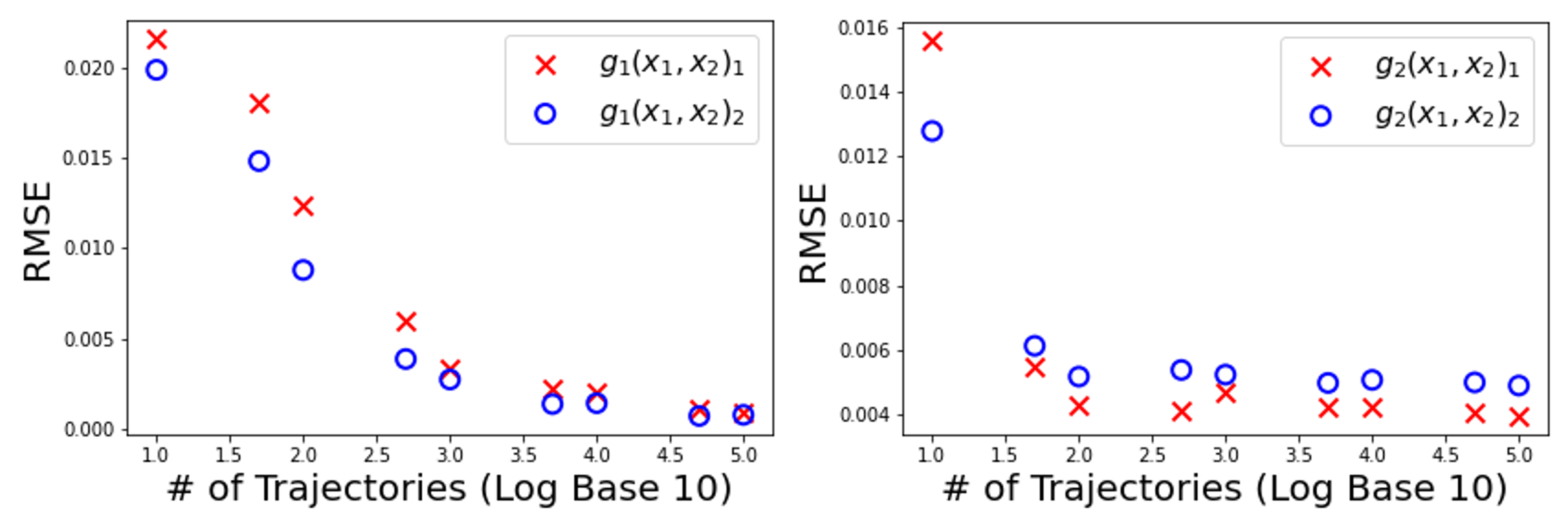}
  \caption{Sampling sensitivity analysis: competitive Lotka-Volterra with a coexistence equilibrium. SPINODE trains the neural networks that approximate $g_1(x_1,x_2)_{1,2}$ and $g_2(x_1,x_2)_{1,2}$ after decreasing the total number of repeated stochastic trajectories used to estimate the moments $m^{(i)}_x(t_{k})$ from data. The root mean square errors (RMSEs) between the learned and actual hidden physics is then calculated. The RMSEs converge around $10^3$ total repeats for $g_1(x_1,x_2)_{1,2}$ and $10^2$ total repeats for $g_2(x_1,x_2)_{1,2}$.
 }
  \label{fgr:lve_numtraj}
\end{figure*}

\begin{figure*}[ht!]
\centering
  \includegraphics[width=0.5\textwidth]{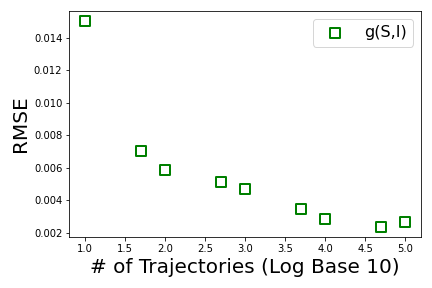}
  \caption{Sampling sensitivity analysis: susceptible-infectious-recovered (SIR) epidemic model. SPINODE trains the neural network that approximates $g(S,I)$ after decreasing the total number of repeated stochastic trajectories used to estimate the moments $m^{(i)}_x(t_{k})$ from data. The root mean square error (RMSE) between the learned and actual hidden physics is then calculated. The RMSE converges around $10^4$ total repeats.
 }
  \label{fgr:sir_numtraj}
\end{figure*}

 Figs. \ref{fgr:lve_numtraj}-\ref{fgr:sir_numtraj} show RMSEs between the learned hidden physics $g(x;\theta^*)$ and the actual system hidden physics $g(x)$ for the Lotka-Volterra and SIR epidemic case studies as a function of the total number of repeated trajectories used to calculate the data-estimated moments $m^{(i)}_x(t_{k})$. As can be seen, the RMSEs converge between $10^3$ and $10^4$ total repeats in both case studies and the RMSEs grow very quickly under $10^2$ total repeats. Figs. \ref{fgr:lve_numtraj}-\ref{fgr:sir_numtraj} highlight that SPINODE's ability to the learn the hidden physics $g(x)$ critically hinges on how accurately moments $m^{(ij)}_x(t_{k})$ can be estimated from data. In this work, moments are estimated from data by repeating (many) stochastic trajectories from identical initial conditions. Section 2.1 discusses how this strategy is not appropriate for systems in which one does not have control over initial conditions, number of replica runs, consistent measurement times, etc. Section 2.1 also suggests potential methods for learning data-estimated moments in such cases.

\par
\begin{figure*}[ht!]
\centering
  \includegraphics[width=\textwidth]{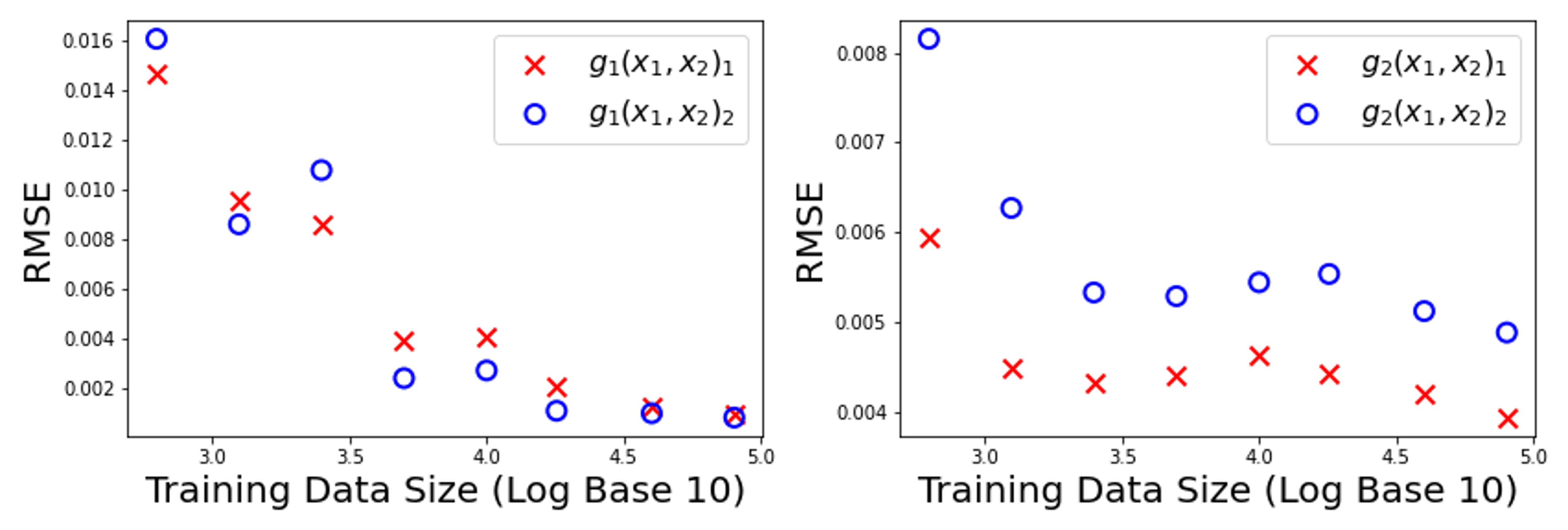}
  \caption{Training data size sensitivity analysis: competitive Lotka-Volterra with a coexistence equilibrium. SPINODE trains the neural networks that approximate $g_1(x_1,x_2)_{1,2}$ and $g_2(x_1,x_2)_{1,2}$ after decreasing the size of the training data (i.e., the total number number of data-estimated moments $m^{(i)}_x(t_{k})$). The root mean square error (RMSE) between the learned and actual hidden physics is then calculated. The RMSEs converge around $2.5 \times 10^4$ total moments for $g_1(x_1,x_2)_{1,2}$ and $5 \times 10^3$ total moments for $g_2(x_1,x_2)_{1,2}$.
 }
  \label{fgr:lve_ts}
\end{figure*}

\begin{figure*}[ht!]
\centering
  \includegraphics[width=0.5\textwidth]{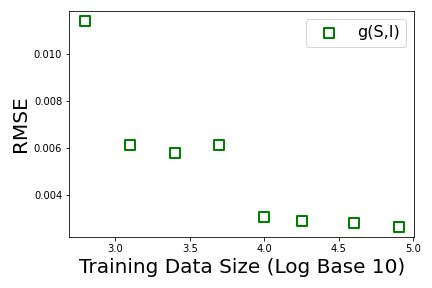}
  \caption{Training data size sensitivity analysis: susceptible-infectious-recovered (SIR) epidemic model. SPINODE trains the neural networks that approximate $g(S,I)$ after decreasing the size of the training data (i.e., the total number number of data-estimated moments $m^{(i)}_x(t_{k})$). The root mean square error (RMSE) between the learned and actual hidden physics is then calculated. The RMSE converges slightly before $10^4$ total moments.
 }
  \label{fgr:sir_ts}
\end{figure*}

 Figs. \ref{fgr:lve_ts}-\ref{fgr:sir_ts} plot the RMSEs between the learned hidden physics $g(x;\theta^*)$ and the actual system hidden physics $g(x)$ for the Lotka-Volterra and SIR epidemic case studies as a function of the total number of data-estimated moments $m^{(i)}_x(t_{k})$ used to train $g(x;\theta)$. In this case, each moment $m^{(i)}_x(t_{k})$ is estimated using $10^5$ total repeated trajectories -- only the total number of moments used to train $g(x;\theta$) is varied. Both figures show that more training data can lead to a more accurate recovery of the hidden physics. The amount of training data required for the RMSEs to converge depends on a combination of the complexity of $g(x)$ and the ``informativeness'' of the loss function used to train $g(x;\theta)$. For example, the behavior of $g(S,I)$ in SIR epidemic case study can be considered  more nonlinear than that of $g_1(x_1,x_2)_1$ and $g_1(x_1,x_2)_2$ in the Lotka-Volterra case study, which is more nonlinear still than that of $g_2(x_1,x_2)_1$ and $g_2(x_1,x_2)_2$ in the Lotka-Volterra case study. Correspondingly, the RMSEs of $g_2(x_1,x_2)_1$ and $g_2(x_1,x_2)_2$ converge after fewer total data points than the other hidden physics terms. Despite the more nonlinear behavior of $g(S,I)$, however, its RMSE converges earlier than the RMSEs of $g_1(x_1,x_2)_1$ and $g_1(x_1,x_2)_2$. We note that the cost function used to train $g(S,I)$ is more ``informative'' than the cost functions used to train $g_1(x_1,x_2)$ and $g_2(x_1,x_2)$ -- compare Eq. \eqref{eqn:simpler_cost} to Eq. \eqref{eqn:utCOST_3}) -- as Eq. \eqref{eqn:utCOST_3} contains added information from multiple known ``physical'' terms in Eq. \eqref{eqn:sirDYN}. Overall, the general notion that more training data can lead to higher-performing neural network models is expected \cite{hinton2012neural}. However, the fact that $g(S,I)$'s RMSE seems to converge at fewer total data points suggests the previously reported observation \cite{raissi2019physics} that incorporating more physics into the cost function can reduce data requirements for training neural networks.

\par
\begin{figure*}[ht!]
\centering
  \includegraphics[width=\textwidth]{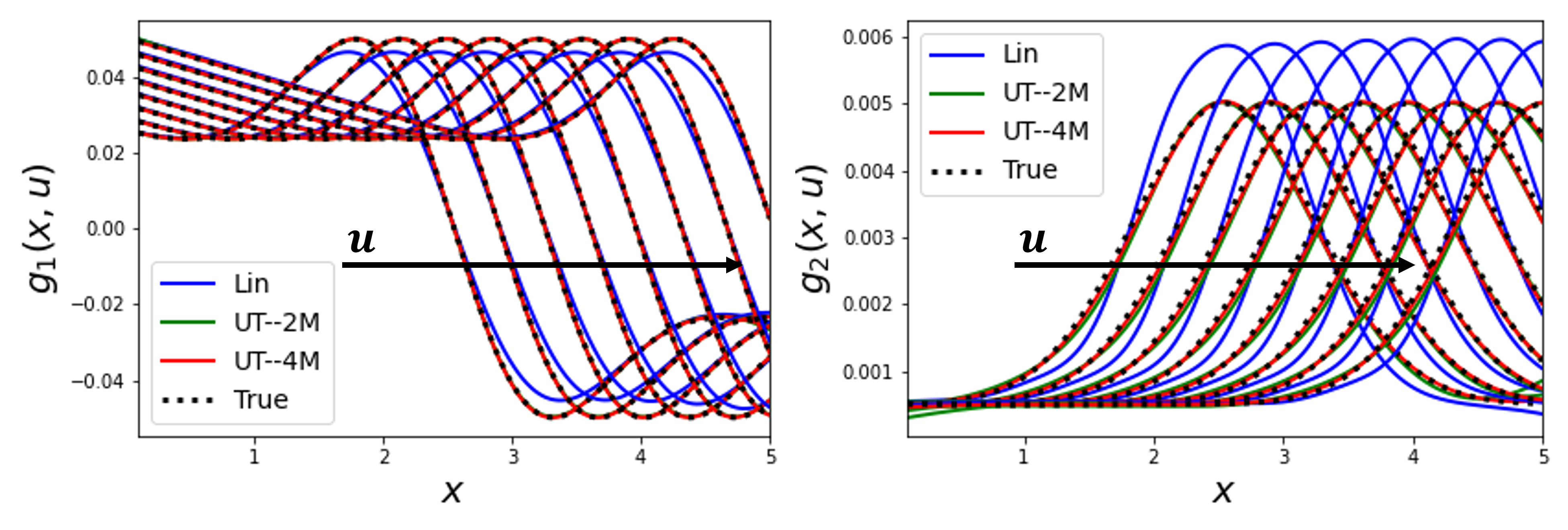}
  \caption{Uncertainty propagation sensitivity analysis: directed colloidal self-assembly with an exogenous input. SPINODE trains neural networks that approximate $g_1(x,u)$ and $g_2(x,u)$ using three different uncertainty propagation methods: linearization (Lin), unscented transform with two moments (UT-2M) and unscented transform with four moments (UT-4M). UT-2M describes the time evolution of the mean and covariance based on the data-estimated mean and covariance at previous time points while UT-4M describes the time evolution of the mean and covariance based on the data-estimated means, covariance, skew, and kurtosis at previous time points. SPINODE with UT-2M and UT-4M significantly outperforms SPINODE with linearization, while SPINODE with UT-4M slightly outperforms SPINODE with UT-2M for learning $g_2(x,u)$.
 }
  \label{fgr:csa_up}
\end{figure*}

 We next use the colloidal self-assembly case study to investigate SPINODE's sensitivity to the chosen uncertainty propagation method. Fig. \ref{fgr:csa_up} shows SPINODE's reconstruction of the hidden physics $g_1(x,u)$ and $g_2(x,u)$ when propagating stochasticity via linearization \cite{paulson2017input} and two methods based on unscented transform -- UT-2M and UT-4M. UT-2M, which is explained in detail in Sections 2.2 and 2.4, describes the time evolution of the mean and covariance based on the data-estimated mean and covariance at previous time points. UT-4M, which can be viewed as an extension of UT-2M based on the work in \cite{ponomareva2010new}, describes the time evolution of the mean and covariance based on the data-estimated means, covariance, skew, and kurtosis at previous time points. SPINODE with both UT methods significantly outperforms SPINODE with linearization. This performance discrepancy indicates that the UT methods propagate stochasticity through Eq. \eqref{eqn:csaDYN} much more accurately than the linearization method does. While SPINODE with UT-2M and UT-4M learn $g_1(x,u)$ with near identical accuracy, the RMSE of SPINODE with UT-4M's recovery of $g_2(x,u)$ is marginally lower than the RMSE of SPINODE with UT-2M's recovery of $g_2(x,u)$ (i.e., $8.64 \times 10^{-5}$ vs. $4.67 \times 10^{-5}$). UT-4M thus leads to a more accurate prediction of the time evolution of the covariance than UT-2M does, as only the covariance is used to train $g_2(x,u)$ (see Eq. \eqref{eqn:simpler_cost}). The latter point supports our earlier remark that SPINODE's ability to incorporate higher moments can make SPINODE well-suited for learning $g(x)$ when $f(x,(g(x))$ and $h(x,(g(x))$ are highly nonlinear and the distribution of $x$ is non-Gaussian as a result. We note that although kernel density estimations in Fig. \ref{fgr:csa_distcomp} appear fairly Gaussian, the relatively minor skews and kurtoses of the distributions of $x(t)$ are still large enough to affect the uncertainty propagation.

\par
\begin{figure*}[ht!]
\centering
  \includegraphics[width=\textwidth]{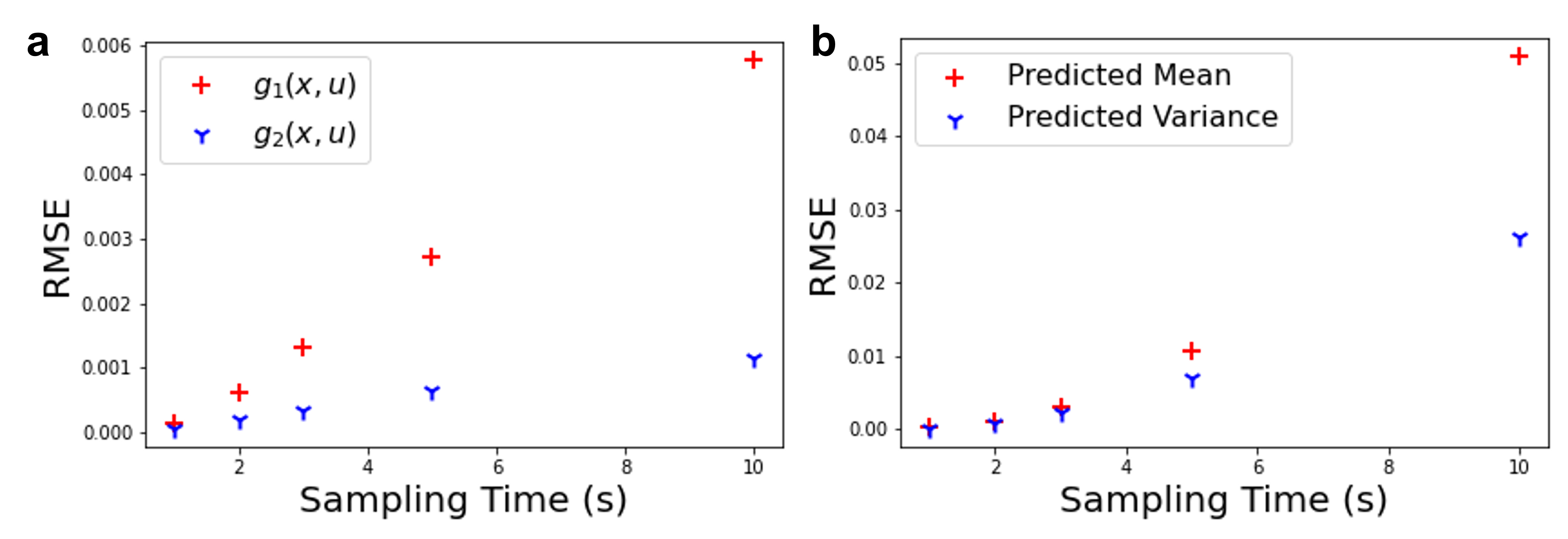}
  \caption{Sampling time sensitivity analysis: directed colloidal self-assembly with an exogenous input. (a) SPINODE trains neural networks that approximate $g_1(x,u)$ and $g_2(x,u)$ using the UT-4M uncertainty propagation method with different sampling times. The root mean square error (RMSE) between the learned and actual hidden physics is then calculated. (b) UT-4M is used to propagate stochasticity through the true dynamics (i.e., Eq. \eqref{eqn:csaDYN}) for each of the data-estimated moments in the training data set at different sampling times. The RMSEs between the predicted and the data-estimated means and covariances at the next sampling time are then calculated. The errors in reconstructing the hidden physics and predicting the mean and covariance grow nearly linearly with the sampling time.
 }
  \label{fgr:csa_st}
\end{figure*}

 We further investigate SPINODE's sensitivity to uncertainty propagation by extending the sampling times at which data-estimated moments are collected. All previous results for the colloidal self-assembly case study used a sampling time of 1 second. Fig. \ref{fgr:csa_st}a plots the RMSEs of $g_1(x,u)$ and $g_2(x,u)$ (when UT-4M is implemented for uncertainty propagation) as a function of sampling time. The RMSE increases nearly linearly with sampling time. Fig \ref{fgr:csa_st}b shows that the prediction errors for the mean and covariance also increase nearly linearly with sampling time. It is thus reasonable to suggest that SPINODE's sensitivity to sampling time in the colloidal self-assembly case study can be attributed to the sensitivity of the uncertainty propagation method to the sampling time.

\par Above all else, Figs. \ref{fgr:csa_up}--\ref{fgr:csa_st} demonstrate SPINODE's sensitivity to the choice of uncertainty propagation method. Section 1 discusses how SPINODE can in principle accommodate any uncertainty propagation method. As a result, the uncertainty propagation method should be viewed as a ``hyper-parameter'' within SPINODE. We finally discuss the choice of the ODE solver within SPINODE. Because the case study simulation data was generated via an Euler-Maruyama discretization, an Euler ODE solver within SPINODE yielded the most accurate reconstructions of the hidden physics. The current implementation of SPINODE \cite{OLeary2022}, however, includes other advanced, even adaptive time-step solvers that have been shown to accurately integrate high-dimensional, stiff, and nonlinear ODEs \cite{chen2018neural}. The choice of ODE solver should thus also be viewed as a ``hyper-parameter'' within SPINODE. In fact, known hyper-parameter optimization strategies such as Bayesian Optimization \cite{snoek2012practical} can be used to determine the ``best'' ODE solver to use during training.

\section{Conclusions and Future Work}

\par We proposed a flexible and scalable framework based on the notions of neural ordinary differential equations, physics-informed neural networks, and moment-matching for training deep neural networks to learn constitutive equations that represent hidden physics within stochastic differential equations. We demonstrated the proposed stochastic physics-informed neural ordinary differential equation framework on three benchmark \textit{in-silico} case studies from the literature. We analyzed the performance of the proposed framework in terms of its repeatability, sensitivity to weight initialization and training/validation/testing set allocation, total number of data points, total number of repeated trajectories, uncertainty propagation method, and sampling time. We showed the framework's scalability by learning highly nonlinear hidden physics within multidimensional stochastic differential equations with multiplicative noise. We illustrated the framework's flexibility by (i) learning both general drift and diffusion coefficients (with or without an exogenous input) and specific unknown functions within stochastic differential equations for different systems and (ii) demonstrating that key aspects of the framework (e.g., the choice of uncertainty propagation method) can be easily and independently adjusted. An open challenge is the fact that a large number of repeated state trajectories are required to accurately learn hidden physics. We will focus future work on learning probability distributions directly from data instead of estimating moments from repeated stochastic trajectories from identical initial conditions. To this end, we will explore variational autoencoders \cite{kingma2013auto, an2015variational, doersch2016tutorial}, generative adversarial networks \cite{yang2019adversarial, goodfellow2014generative, zoufal2019quantum, mescheder2017adversarial}, and energy-based models \cite{lecun2006tutorial, kim2016deep, gustafsson2020energy}. Other open challenges include optimizing the uncertainty propagation method and choice of ODE solver during neural network training. We will explore optimizing these ``hyper-parameter'' choices using methods based on Bayesian optimization \cite{snoek2012practical}. We will finally explore updating the current implementation of the neural ODE framework based on \cite{chen2018neural} with more recent and advanced neural ODE framework implementations (e.g., \cite{rackauckas2020universal}).

\section{Acknowledgment}

This work is in part supported by the National Science Foundation under Grant 2112754.

 \bibliographystyle{elsarticle-num} 
 \bibliography{cas-refs}





\end{document}